\documentclass[preprints,article,accept,moreauthors,pdftex]{Definitions/mdpi} 

\firstpage{1} 
\pubvolume{1}
\issuenum{1}
\articlenumber{0}
\pubyear{2021}
\copyrightyear{2021}
\externaleditor{Academic Editor: Chen Chen}
\datereceived{} 
\dateaccepted{} 
\datepublished{} 
\hreflink{https://doi.org/} 

 \usepackage[labelformat=simple]{subcaption}

\DeclareCaptionLabelFormat{subcaptionlabel}{\normalfont(\textbf{#2}\normalfont)}
\graphicspath{{img/}}

\Title{Deep HDR Hallucination for Inverse Tone Mapping}

\TitleCitation{Deep HDR Hallucination for Inverse Tone Mapping}

\Author{Demetris Marnerides 
 $^{1}$, Thomas Bashford-Rogers $^{2}$ and Kurt Debattista $^{1,}$*}

\AuthorNames{Demetris Marnerides, Thomas Bashford-Rogers and Kurt Debattista}

\AuthorCitation{Marnerides, D.; Bashford-Rogers, T.; Debattista, K.}

\address{%
$^{1}$ \quad WMG, University 
 of Warwick
, Coventry,  CV4 7AL 
, UK
; Demetris.Marnerides@warwick.ac.uk\\
$^{2}$ \quad Department of Computer Science and Creative Technologies, University of the West of England, Bristol, BS16~1GY, UK; Tom.Bashford-Rogers@uwe.ac.uk\\
}

\corres{Correspondence: K.Debattista@warwick.ac.uk}

\abstract{Inverse Tone Mapping (ITM) methods attempt to reconstruct High Dynamic Range (HDR) information from Low Dynamic Range (LDR) image content. The dynamic range of well-exposed areas must be expanded and any missing information due to over/under-exposure must be recovered (hallucinated). The majority of methods focus on the former and are relatively successful, while most attempts on the latter are not of sufficient quality, even ones based on Convolutional Neural Networks (CNNs). A major factor for the reduced inpainting quality in some works is the choice of loss function. Work based on Generative Adversarial Networks (GANs) shows promising results for image synthesis and LDR inpainting, suggesting that GAN losses can improve inverse tone mapping results. This work presents a GAN-based method that hallucinates missing information from badly exposed areas in LDR images and compares its efficacy with alternative variations. The proposed method is quantitatively competitive with state-of-the-art inverse tone mapping methods, providing good dynamic range expansion for well-exposed areas and plausible hallucinations for saturated and under-exposed areas. A density-based normalisation method, targeted for HDR content, is also proposed, as well as an HDR data augmentation method targeted for HDR hallucination.}


\begin{document}
\section{Introduction}\label{sec:hall:intro}

High Dynamic Range (HDR) imaging~\cite{banterle2017newbook} permits the manipulation of content with a high dynamic range of luminance, unlike traditional imaging typically called standard or Low Dynamic Range (LDR). While HDR has recently become more common, it is not the exclusive mode of imaging yet, and~a significant amount of content exists which is still LDR. Inverse tone mapping or dynamic range expansion~\cite{banterle06itm} is the process of converting LDR images to HDR to permit legacy LDR content to be viewed in~HDR. 

The problem of inverse tone mapping is essentially one of information recovery, with~arguably its hardest part being the inpainting/hallucination of over-exposed and under-exposed regions. In~these regions, there is not sufficient information in the surrounding pixels of the LDR input for interpolation, compared to regions that are, for~example, LDR only due to quantisation and thus preserve some colour and structure information. In~addition, inpainting is content and context specific, while other aspects of dynamic range expansion are not very dependent on semantic aspects of the~scene.

Deep neural networks are powerful function approximators and CNNs have been shown to be good candidates to target the inpainting problem~\cite{iizuka2017completion, pathakcontext}.  However, in~this case, it is not only the network architecture that needs to be carefully selected, but~the optimisation loss is of paramount importance as well. For~example, in~other inverse tone mapping works~\cite{marnerides2018exp, marnerides2020spectrally, eilertsen2017cnn, endo2017drtmo}  
the CNNs presented are trained with standard regression losses (\(L_1\), \(L_2\) and/or cosine similarity) and yield good results when considering well exposed areas, but~they either do not reconstruct any missing information of saturated areas, or~provide some reconstruction in these areas but of not sufficient~quality.

In this work, a~GAN-based solution to the Inverse Tone Mapping problem is proposed, providing simultaneous dynamic range expansion and hallucination. GANs can provide a solution to the hallucination problem through the introduction of a learnable loss function, i.e.,~the discriminator network, which being a CNN itself is powerful enough to capture multiple modalities. The~method is end-to-end, producing a linear HDR image from a single LDR input. To~handle the extreme nature of the exponentially distributed linear HDR pixel values of natural images a novel HDR pixel distribution transform is proposed as well, mapping the HDR pixels to a more manageable Gaussian~distribution.

In summary, the~primary contributions of this work are:

\begin{itemize}
    \item The use of GANs for the reconstruction of missing information in over-exposed and under-exposed areas of LDR images.
    \item An HDR pixel distribution adaptation method, for~normalising HDR content, which is exponentially skewed towards zero.
    \item A data augmentation method for the use of more readily available LDR datasets to train CNNs for HDR hallucination and inverse tone mapping.
    \item Comparisons with variants of the proposed GAN-based method using different network architectures and modules.
\end{itemize}

\section{Background and Related~Work}\label{sec:background}
Inverse Tone Mapping Operators (ITMOs), attempt to generate HDR from LDR content. They can generally be expressed as:
\begin{equation}
\tilde{I}_{\text{HDR}} = f_{\text{ITM}}(I_{\text{LDR}}), \text{ where } f_{\text{ITM}}:  [0,255] \to \mathbb{R}^{+}
\label{eqn:eo}
\end{equation}

\noindent where the predicted HDR image is denoted as \(\tilde{I}_{\text{HDR}}\) and the ITMO as \(f_{\text{ITM}}\)\@. If~we consider a badly exposed part of an image, the~missing content from the original scene is completely lost and can take multiple reasonably valid shapes and forms, e.g.,~the over-exposed part of a road may have contained parts of a sidewalk or alternatively a pedestrian. We can therefore say that the ITM problem is an ill-posed~one.

The main procedure followed by ITMOs that are non-learning based is composed of the following steps~\cite{banterle2017newbook}:

\begin{enumerate}
    \item Linearisation: Apply inverse CRF and remove gamma.
    \item Expansion: Well-exposed areas are expanded in dynamic range. The~operation can be local or global in luminance space.
    \item Hallucination: Reconstruct missing information in badly-exposed areas. Not all methods have this step.
    \item Artefact removal: Remove or reduce quantisation or compression artefacts.
    \item Colour correction: Correct colours that have been altered due to saturation of only a single or two channels from the RGB image.
\end{enumerate}

The main distinction between ITMOs is whether the expansion is local or global. Global methods apply the same luminance expansion function per pixel, that has the same form independently of pixel location. The~method by Landis~\cite{landis02}, which was one of the earlier ones, expands the range using power functions, ultimately to have the content displayed on an HDR monitor. Aky\"{u}z~et~al.~\cite{akyuz2007hdr} expand single LDR exposures using linear transformations and gamma correction, while Masia~et~al.~\cite{masia09,masia2017dynamic} propose a multi-linear model for the estimation of the gamma~value.

Local methods use analytical functions to expand the range, which unlike global methods, depend on local image neighbourhoods. Banterle~et~al.~\cite{banterle06itm}, expand the luminance range by applying the inverse of the any invertible Tone Mapping Operator (TMO) and use a smooth low frequency expand map to interpolate between the expanded and the LDR image luminance. Rempel~et~al.~\cite{rempel06ldr2hdr} also use an expand map, termed brightness enhancement function (BEF). However, this is computed through the use of a Gaussian filter in conjunction with an edge-stopping function to maintain contrast.  Kovaleski and Oliviera~\cite{kovaleski2014} extend the work of Rempel~et~al.~\cite{rempel06ldr2hdr} by changing the estimation of the BEF\@. Subsequently, Huo~et~al.~\cite{huo2014} further extend these methods by removing the thresholding used by Kovaleski and~Oliviera.

A method that includes inpainting of missing content is proposed by Wang~et~al.~\cite{wang2007hdrh}.  It is partially user-based, where the user adds the missing information on the expanded LDR image. Kuo~et~al.~\cite{kuo2014automatic} attempt to automatically produce inpainted content, using textures that are similar to the surroundings of the over-exposed areas. It is, however, limited in the type of content, as~the neighbouring textures must be matching and must be of only one single~type. 

Recently, a~variety of deep learning-based ITMOs have been introduced.
Zhang and Lalonde~\cite{zhang2017learninghdr} predict HDR environment maps from
captured LDR panoramas, while Eilertsen~et~al.~\cite{eilertsen2017cnn}, predict content for small over-saturated areas of the LDR input using a CNN, and~use an inverse CRF to linearise the rest of the input. Endo~et~al.~\cite{endo2017drtmo} predict multiple exposures from a single LDR input using a CNN and fuse them using a standard merging algorithm. Marnerides~et~al.~\cite{marnerides2018exp} use a multi-branch architecture for predicting HDR from LDR content in an end-to-end fashion.  
The spectrally consistent UNet architecture, termed GUNet~\cite{marnerides2020spectrally}, has also been used to for end-to-end HDR image reconstruction. Lee~et~al.~\cite{lee2018deepchain} use a dilated CNN to
infer multiple exposures from a single LDR image sequentially, using a
chain-like structure, which are then fused. Han~et~al.~\cite{han2020neuromorphic} propose using a neuromorphic camera in conjuction with a CNN  to help guide HDR reconstruction from a single LDR capture, while Sun~et~al.~\cite{Sun_2020_CVPR} use a Diffractive Optical Element in conjunction with a CNN and Liu~et~al.~\cite{Liu_2020_CVPR} invert the capturing pipeline and use a CNN in conjunction for the same purposes. {Sharif~et~al.~\cite{sharif2021hdr} use a two stage network for reconstructing HDR images focusing on de-quantisation, while Chen~et~al.~\cite{chen2021hdrunet} combine inverse tone mapping with denoising.}
Concurrently to this work, Santos~et~al.~\cite{santos2020ldrhdr} follow a feature-masking methodology along with a perceptual loss to simultaneously expand well-exposed areas and hallucinate badly exposed areas. In~contrast to this work, the~author's do not follow a GAN-based approach and the CNN architectures are more complicated, implementing masked features at various~layers.

\section{Method}\label{sec:method}

The proposed Deep HDR Hallucination (DHH) method attempts to expand the range of well-exposed areas and simultaneously fill in missing information in under/over-exposed areas in an end-to-end fashion. First, an~overview of the method is given, followed by the presentation of the GAN-based loss function and the proposed HDR pixel distribution transformation for HDR content~normalisation. 

\subsection{Network~Architecture}

\begin{figure}[H]
    \includegraphics[width=0.8\linewidth]{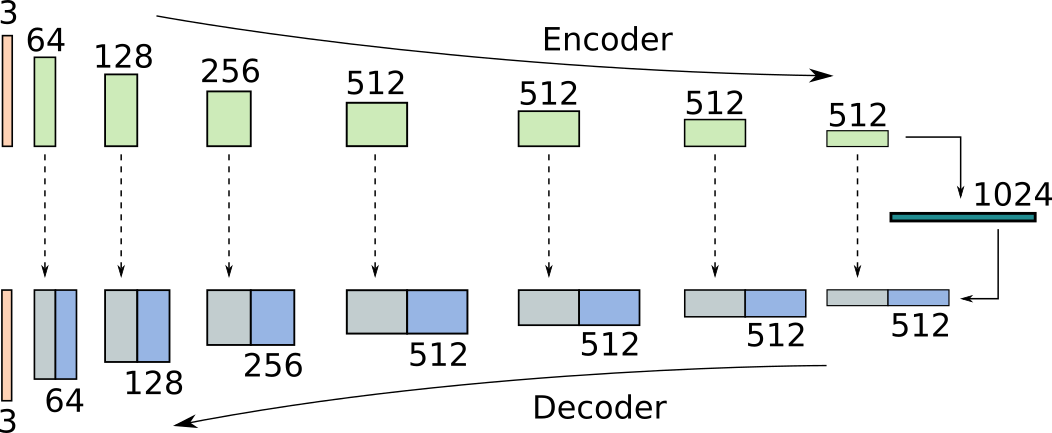}
    \caption{{Network architecture based on the UNet architecture~\cite{ronneberger2015unet} with skip connections.}}\label{fig:network}
\end{figure}

The proposed network is based on the UNet architecture~\cite{ronneberger2015unet}, which has been successfully used previously for image translation tasks using GANs~\cite{isola2016pix2pix, wangpix2pixhd}.
The encoder and decoder, downsample and upsample eight times, forming a total of 9 levels of feature resolutions with feature sizes 3-64-128-256-512-512-512-512-1024, similarly to Isola~et~al.~\cite{isola2016pix2pix}.
At each level, the~encoder uses a residual block as proposed by He et el~\cite{he2016residualmappings} formed of two $4 \times 4$ convolutional layers along with Instance Normalisation and the SELU~\cite{klambauer2017selu} activation, while the decoder fuses the encoder features with the upsampled features using a single convolution. Downsampling and upsampling is performed using residual bilinear modules. {A diagram of the architecture is shown in Figure~\ref{fig:network}.}

\begin{figure}[H]
    \includegraphics[width=0.8\linewidth]{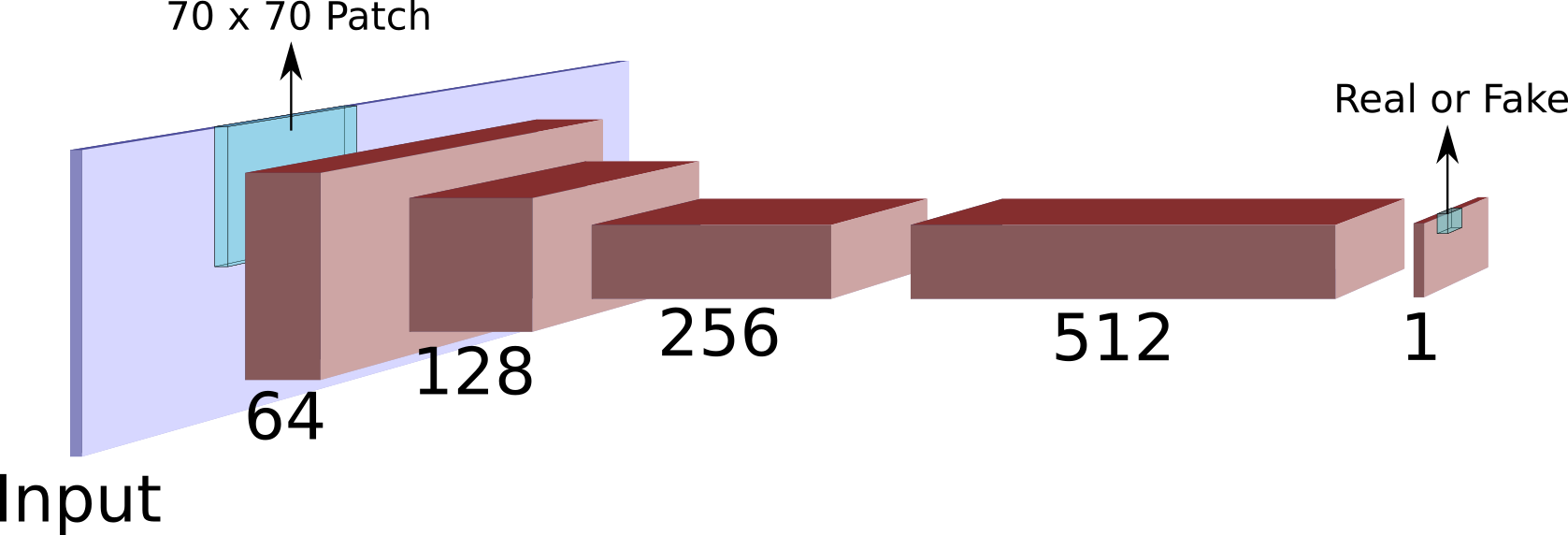}
    \caption{{Diagram of the patch-based discriminator network.}}\label{fig:discriminator}
\end{figure}

{The discriminator network is similar to the PatchGAN discriminator used by \linebreak  Isola~et~al.~\cite{isola2016pix2pix}} for image translation
It consists of five layers with 64-128-256-512-1 features, respectively, {as shown in Figure~\ref{fig:discriminator}}. The~first three layers reduce the spatial dimensionality using bilinear downsampling with a factor of 2. The~final layer output consists of a single channel of size \(32 \times 32\) (for an input of \(256 \times 256\)). Each pixel of the output is a discriminatory measure of an input patch, signalling whether the input patch is from the data distribution or if it's a generator sample. The~discriminator uses the Leaky ReLU activation~\cite{maasleaky}, with~\(\lambda = 0.2\).
\vspace{-6pt}

\subsection{GAN~Objective}

The original optimisation objective for GANs is often found to be unstable~\cite{karrasprogan, salimansgan, kodaliconvergence} and many variations of the loss have been proposed attempting to provide stable alternatives. The~loss function DHH adopts is the Geometric GAN method~\cite{limgeogan} which utilises the hinge loss, commonly used for binary classification:
\begin{equation}
    L_{\text{hinge}}(x, y) = \max(0, 1 - xy)
\end{equation}

\noindent where \(y\) is the target label (\(0\) or \(1\)) and x is the classifier prediction (as a probability). Applying it in the context of GAN, given a generator network, \(G\), and~a discriminator network, \(D\), the~GAN losses are given by:
\begin{equation}\label{eq:dloss}
    L_{D,\text{GAN}} =\text{ } \mathbb{E}_{I_{\text{HDR}}} \left[\max(0, 1 - D(I_{\text{HDR}}))\right]  + \mathbb{E}_{I_{\text{LDR}}}\left[\max(0, 1 + D(G(I_{\text{LDR}})))\right]
\end{equation}
\begin{equation}
    L_{G,\text{GAN}} = \text{ } -\mathbb{E}_{I_{\text{LDR}}}\left[D(G(I_{\text{LDR}}))\right]
\end{equation}

\noindent where \(I_{\text{HDR}}\) are target HDR images, corresponding to LDR inputs, \(I_{\text{LDR}}\) from the dataset. The~final predictions, \(\tilde{I}_{\text{HDR}}\) are the generator network outputs:
\begin{equation}
\tilde{I}_{\text{HDR}} = G(I_{\text{LDR}})
\end{equation}

The hinge loss objective was found to provide stable training for GANs in imaging problems~\cite{zhangsagan} including the large scale {``BigGAN''} study~\cite{brockbiggan} for image generation. In~addition, the~generator is optimising two additional losses which help stabilise the training and improve the quality of the results, one being the perceptual loss~\cite{johnsonperceptual} and the other being the feature matching loss~\cite{wangpix2pixhd}, which is similar to the perceptual loss. 
The optimisation loss for the generator network is given by:
\begin{equation}\label{eq:total_gan}
    L_{G,\text{total}} = L_{G,\text{GAN}} + L_{\text{perc}}(\tilde{I}_{\text{HDR}},I_{\text{HDR}})
        + L_{\text{fm}}(\tilde{I}_{\text{HDR}},I_{\text{HDR}}).
\end{equation}

{The perceptual loss, $L_{\text{perc}}$, is the same as the one used in~\cite{johnsonperceptual} . It is based on a VGG architecture~\cite{DBLP:journals/corr/SimonyanZ14a}, trained on the ImageNet dataset~\cite{deng2009imagenet}. Perceptual losses can recover improved textures since they take into account pixel inter-correlations locally since they use early layers of the pre-trained network.
The feature matching loss, $L_{\text{fm}}$, is calculated as in~\cite{wangpix2pixhd} in the same way that the perceptual loss is calculated on the VGG features, but~on the PatchGAN discriminator instead. As~pointed out in~\cite{wangpix2pixhd} it can help generate more consistent structures, which in our case are the the under/over exposed areas.}
The discriminator loss is the same as in Equation~(\ref{eq:dloss}).

\subsection{Data~Augmentation}\label{sec:data_augmentation}

Generative imaging problems require large datasets, of~the order of tens of thousands to millions of samples, along with large models for improved results, as~demonstrated by Brock~et~al.~\cite{brockbiggan}.  Readily available HDR datasets are small in size, for~example, in~the works by Eilertsen~et~al.~\cite{eilertsen2017cnn} and Marnerides~et~al.~\cite{marnerides2018exp} the datasets used are of the order of 1000 HDR images. For~this reason, a~data augmentation method is proposed, which uses a pre-trained GUNet model~\cite{marnerides2020spectrally} for dynamic range expansion. A~large readily available LDR dataset is used which is then expanded using the pre-trained model in order to simulate an HDR dataset.  To~create input-output pairs the following procedure is followed:

\begin{enumerate}
    \item Sample an image, \(I_d\) from the LDR dataset.
    \item Expand the image range using a trained GUNet to predict the HDR image
        \(I_{\text{HDR}}\).  
    \item Crop and resize the image to \(256 \times
        256\) using the approach from Marnerides~et~al.~\cite{marnerides2018exp}.
    \item Use the Culling operator (clipping of top an bottom \(10\%\) of
        values).
\end{enumerate}

The culling operator on the last step discards information from the expanded HDR target image which the hallucination network attempts to reconstruct. The~dataset is sourced from Flickr~\cite{flickr} and consists of 45,166 training images from 17 different categories using the following tags: Cloud, fields, lake, landscape, milkyway, mountain, nature, panorama, plains, scenery, skyline, sky, sunrise, sunset, sun, tree, water.

\subsection{HDR Pixel~Transform}\label{sec:gunet:transform}

\begin{figure}[H]
    \begin{subfigure}[t]{0.4\linewidth}
        \includegraphics[width=1.0\linewidth]{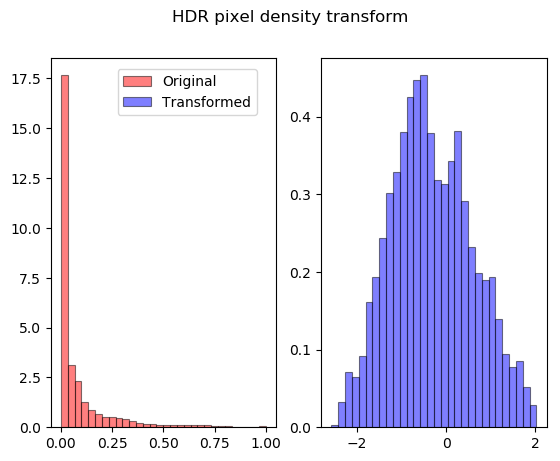}
    \end{subfigure}
    \begin{subfigure}[t]{0.4\linewidth}
        \includegraphics[width=1.0\linewidth]{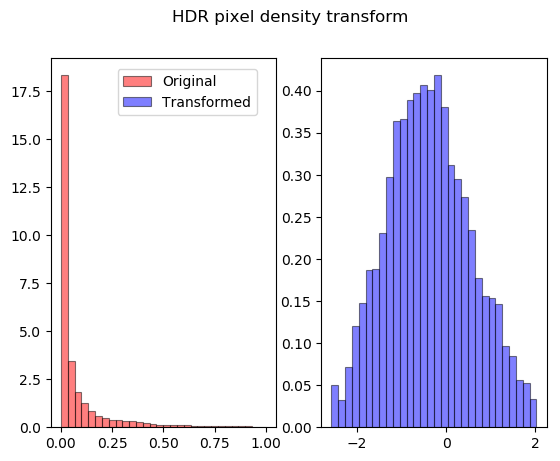}
    \end{subfigure}\\
    \begin{subfigure}[t]{0.4\linewidth}
        \includegraphics[width=1.0\linewidth]{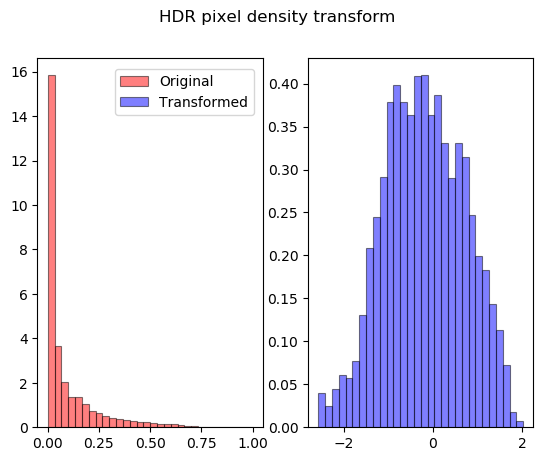}
    \end{subfigure}
    \begin{subfigure}[t]{0.4\linewidth}
        \includegraphics[width=1.0\linewidth]{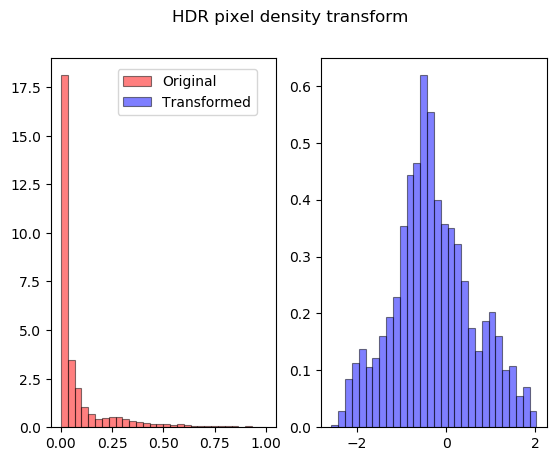}
    \end{subfigure}
    \caption[HDR pixel distributions before and after the density
    transform]{Histograms for HDR pixel distributions before and after the
        density transform. The~four subfigures represent four different samples
        of \(256 \times 256\) HDR image mini-batches of size 32. There is some
    variation observed in the resulting Gaussian distributions, due to the
variation in HDR content and the approximation that HDR pixels follow an
exponential distribution.}\label{fig:pixel_transform}
\end{figure}

The HDR reconstruction loss used in recent CNN-based ITM methods~\cite{marnerides2018exp}, comprise
of an \(L_1\) loss along with a cosine similarity term. This specific loss,
particularly the cosine similarity term is used to accommodate the unique
distribution of HDR image pixel values, shown in red in
Figure~\ref{fig:pixel_transform}. The~cosine term, being magnitude independent, helps improve colour
consistency in the darker areas, where deviations between the RGB channels do not influence the main regression loss (\(L_1\)) greatly thus producing deviations in~chrominance.

However, having normally distributed data is beneficial for training CNNs, hence a variety of methods
attempt to induce normalised outputs from activations~\cite{klambauer2017selu,ioffe2015batchnorm}. Thus, instead of adjusting the training loss function to fit the distribution, the~distribution can be normalised. The~proposed method applies such a transformation on the original HDR pixels for inputting into both the generator and the discriminator. The~blue histograms in Figure~\ref{fig:pixel_transform} show the HDR pixel
distributions after applying a transformation function, such that the resulting
distribution is approximately~Gaussian.

To derive the transformation function it is first noted that the HDR pixels
\(x\) approximately follow an exponential distribution of the form:
\begin{equation}
    x \sim \lambda e^{-\lambda x}
\end{equation}

\noindent with the parameter lambda estimated from the HDR dataset to be \(\lambda \approx
1\).
To transform an exponentially distributed variable \(X\) to become a
Gaussian variable \(Y\) the following equation is used:
\begin{equation}\label{eqn:pix_trans}
    Y = \Phi^{-1} ( F_Y (Y))
\end{equation}

\noindent where \(\Phi\) is the Cumulative Distribution Function (CDF) of the Gaussian
distribution and \(F_Y\) is the CDF of the exponential distribution:
\begin{equation}
    \Phi(x) = \frac{1}{2} \left[ 1 + \text{erf}\left( \frac{x}{\sqrt{2}} \right) \right].
\end{equation}

\section{Results and~Discussion}\label{sec:results}

This section presents both qualitative and quantitative results. For~both, DHH is trained using four Nvidia P100 GPUs for approximately 700,000 iterations using a mini-batch of size 36. The~generator and discriminator updates are performed with the same frequency (alternating every one step). The~learning rates used were 
 \(1e^{-4}\) for the generator and \(4e^{-4}\) for the discriminator, following the Two Timescale Update Rule (TTUR)~\cite{heuselttur} using the Adam optimiser with \(\beta_1 = 0.5\) and \(\beta_2=0.999\). Both networks make use of spectral normalisation~\cite{miyato2018spectral} to help with training stability by controlling their Lipschitz constant via limiting the highest singular value (spectral norm) of the linear transformation weight~matrices.

\subsection{Qualitative}\label{sec:qualitative}

\begin{figure}[H]
    \includegraphics[width=1.0\linewidth]{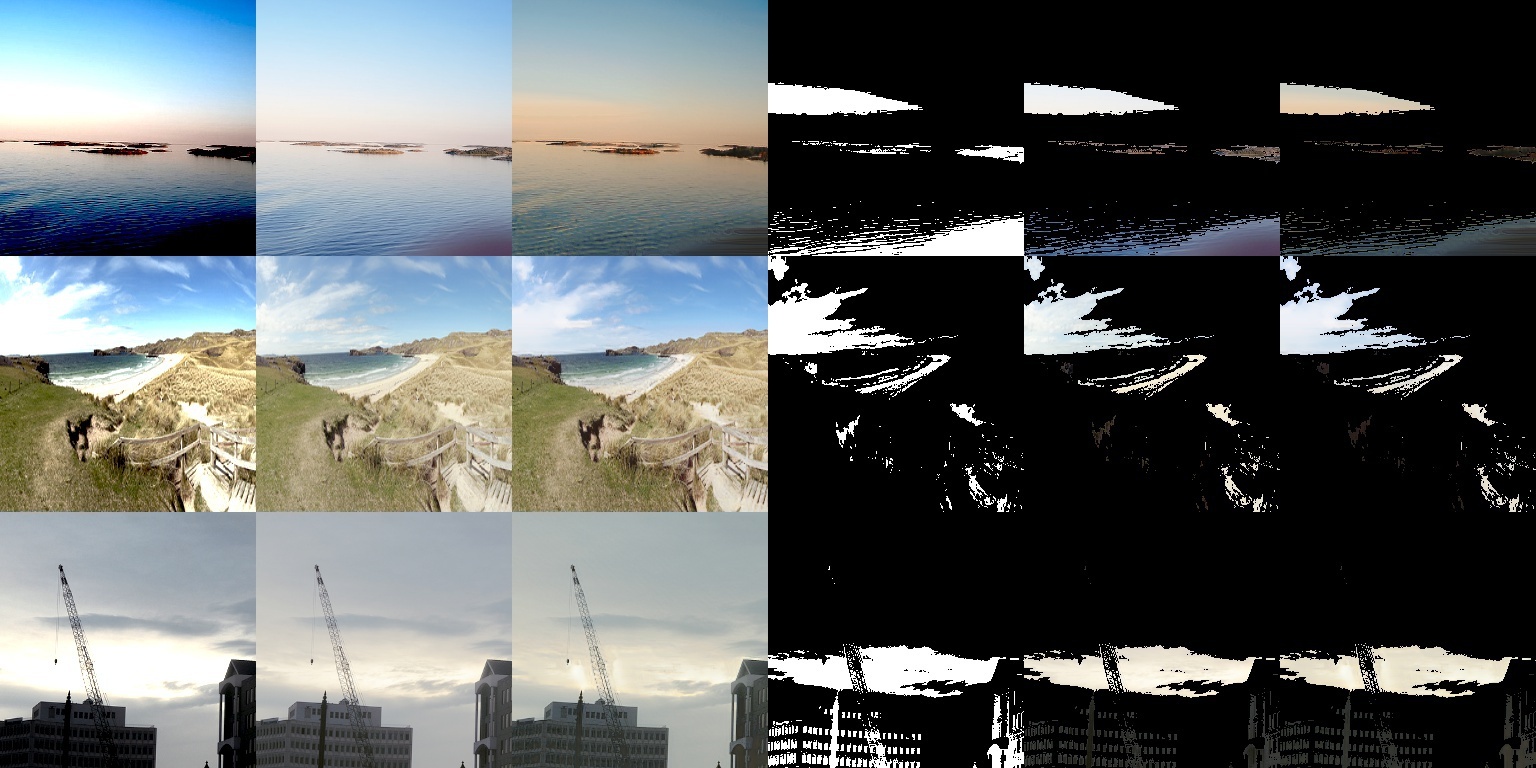}
    \caption{Samples for the DHH method (training). Columns
        1--3 are the input LDR image and corresponding tone mapped HDR target and prediction, respectively.
        Column 4 is a mask showing the fully saturated (\([255,255,255]\)) and fully under-exposed (\([0,0,0]\)) pixels of the input.
        Columns 5 and 6 show the masked target and masked prediction, respectively.}\label{fig:preds}
\end{figure}

Figure~\ref{fig:preds} presents sample predictions during training. The~first column is the input image, while the second and third columns are the tone-mapped target and prediction images, respectively. The~fourth column is the masks of the badly-exposed areas in the input, highlighting the pixels that are completely black or white. The~last two columns highlight the badly exposed areas in the target and the prediction~respectively. 

It is observed that the well-exposed areas of the image are reconstructed well, matching the corresponding areas in the target image. In~the badly exposed areas, textures are reconstructed to match the surroundings.
The hallucinations blend well with the rest of the image and also exhibit texture variations. For~example in the first row, the~sky is reconstructed and varies from blue to light red towards the horizon.  The~second and fourth rows introduce variations in the clouds. However, semantic detail is not reconstructed fully, for~example potentially the shape of the sun in the third~row.

\begin{figure}[H]
{\captionsetup{position=bottom,justification=centering}
    \begin{subfigure}[t]{0.55\linewidth}
        \includegraphics[width=1.0\linewidth]{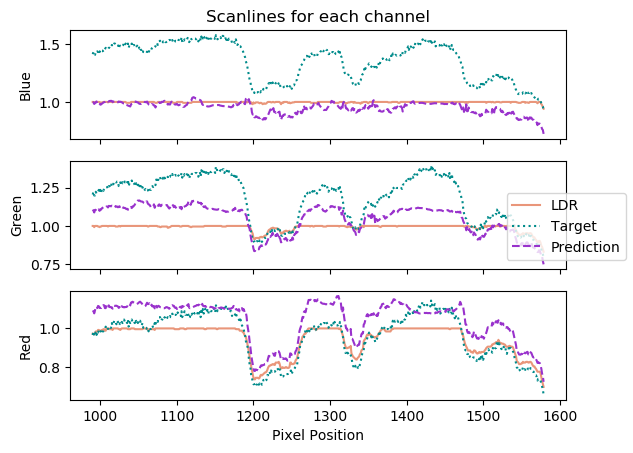}
        \caption{Scan-line~plot}
    \end{subfigure}\hspace{2em}
    
    \par\bigskip
    \begin{subfigure}[t]{0.32\linewidth}
        \includegraphics[width=1.0\linewidth]{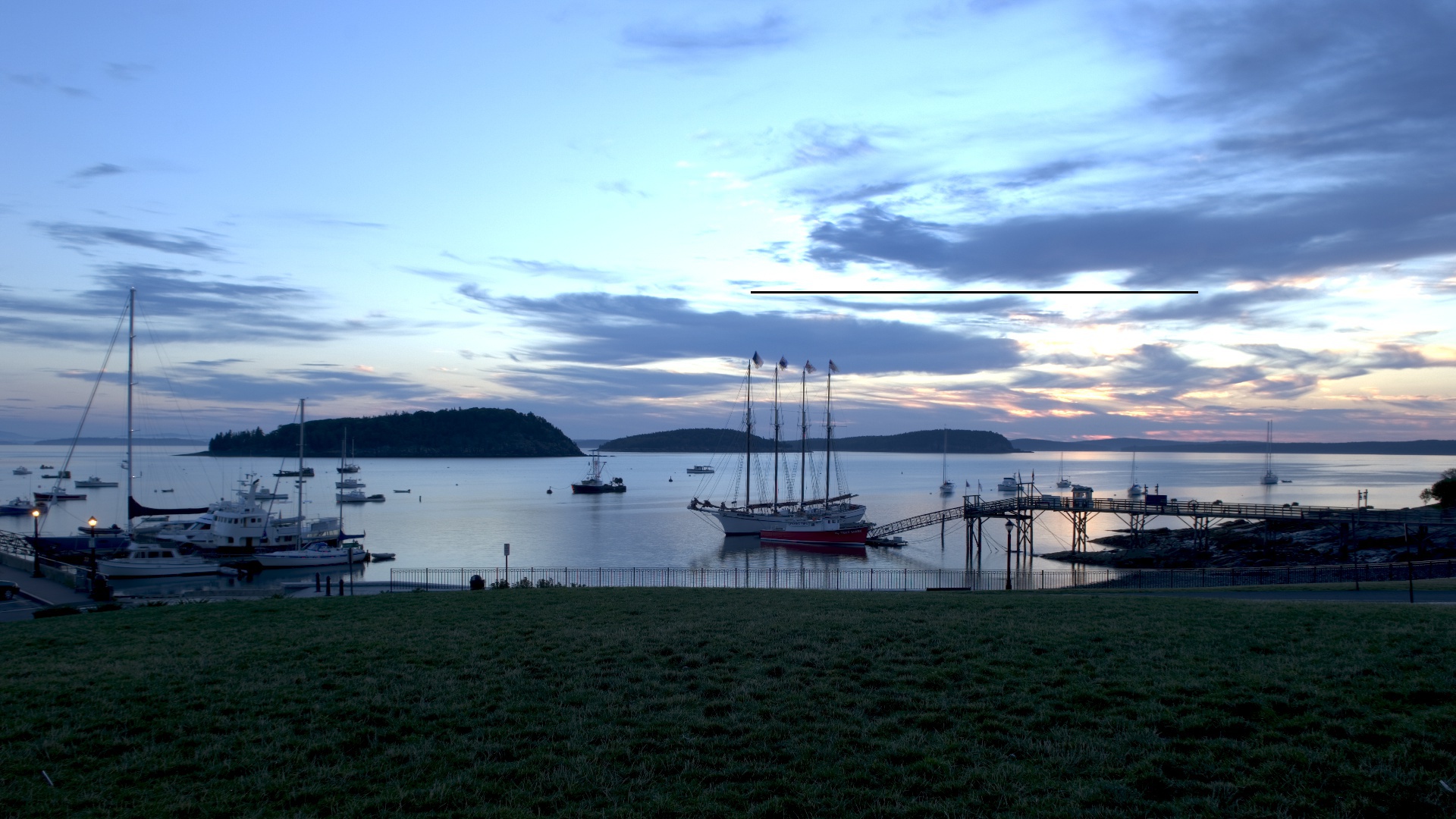}
        \caption{Input~LDR}
    \end{subfigure}
    \begin{subfigure}[t]{0.32\linewidth}
        \includegraphics[width=1.0\linewidth]{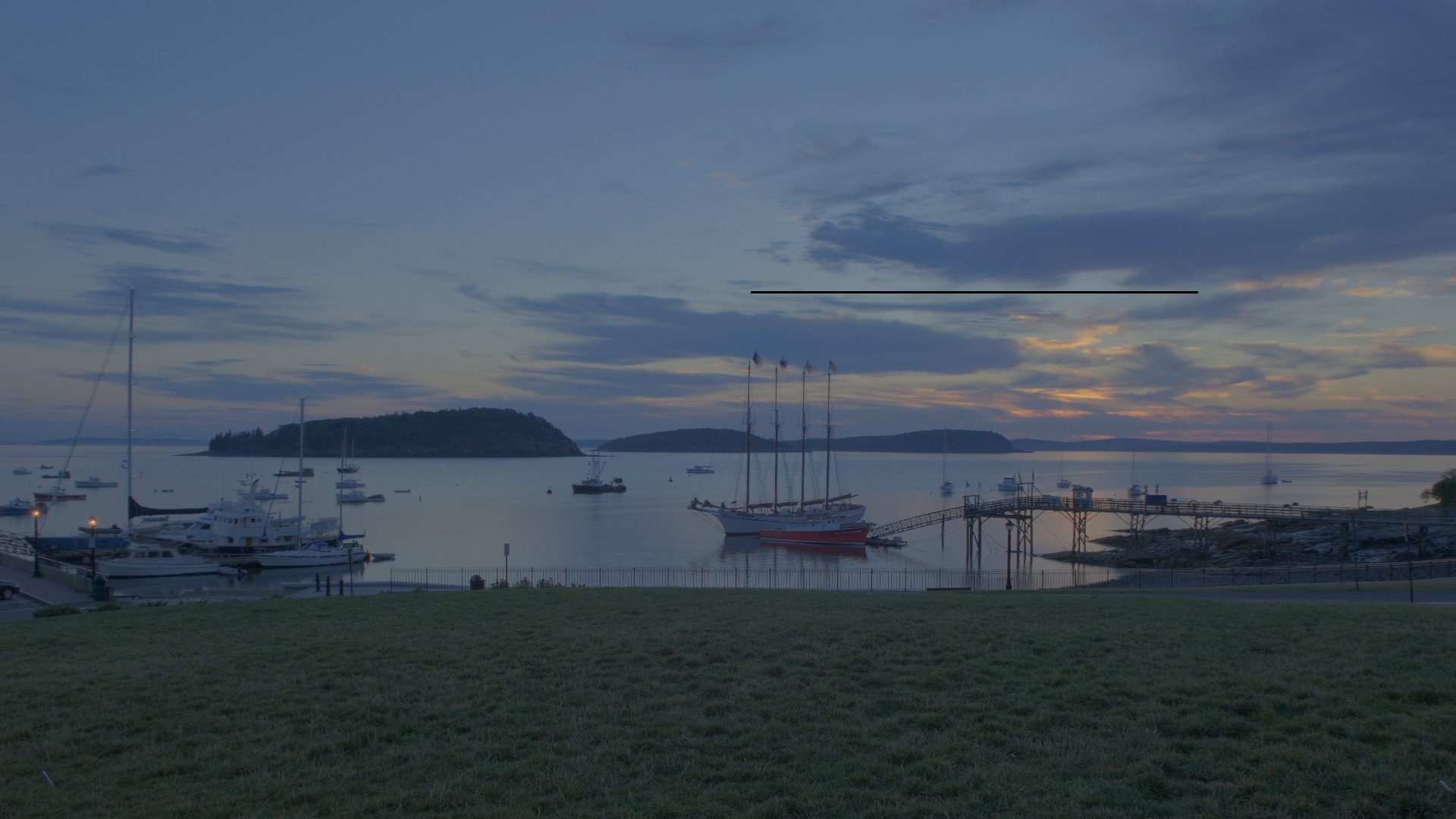}
        \caption{Target~HDR}
    \end{subfigure}
    \begin{subfigure}[t]{0.32\linewidth}
        \includegraphics[width=1.0\linewidth]{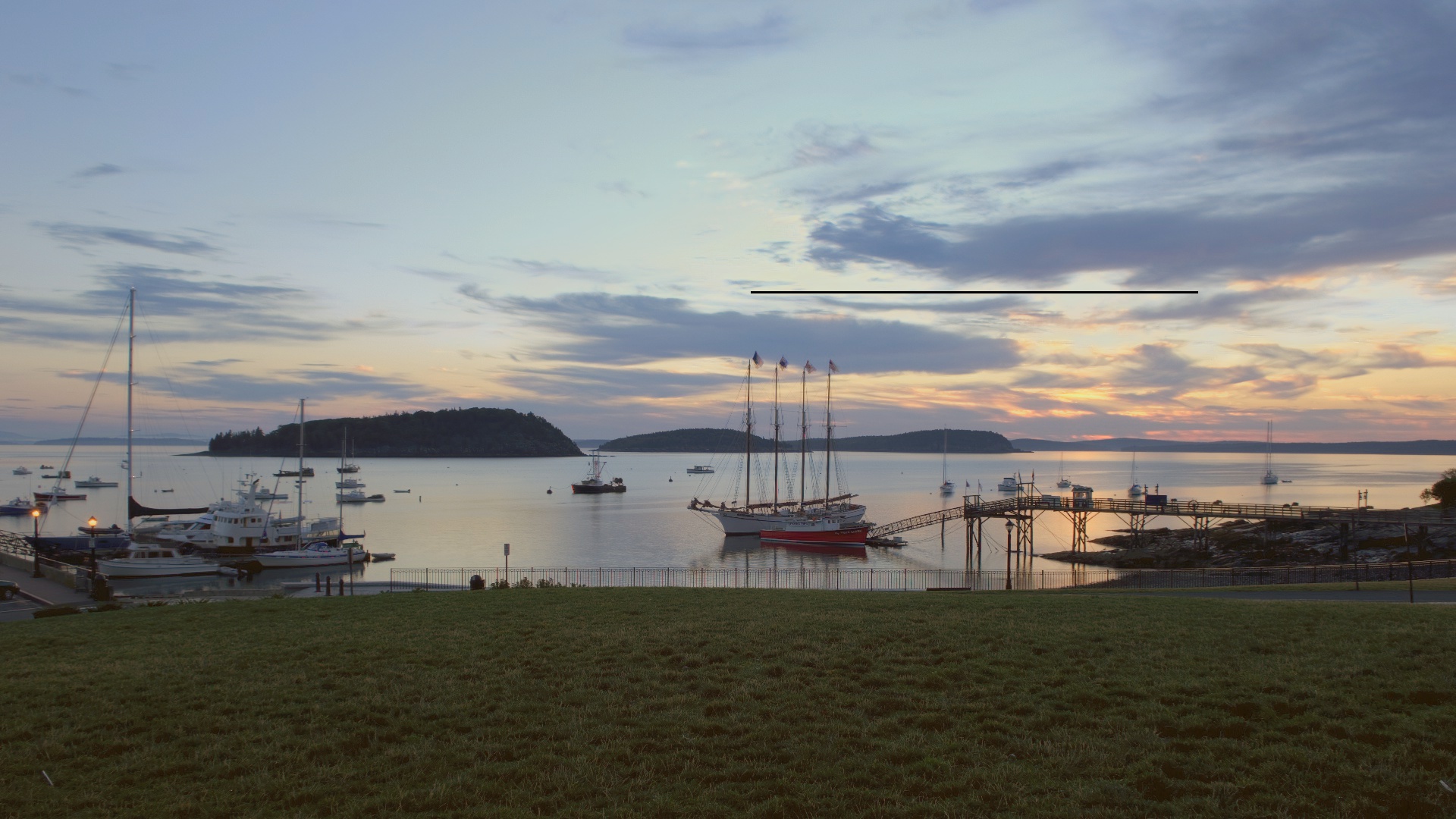}
        \caption{Prediction~HDR}
    \end{subfigure}}
   
    \caption{Scan-line plot for a test-set image prediction using DHH\@. The~saturated areas in
    the LDR are reconstructed well in the prediction and correlate with the
    target HDR\@.}\label{fig:scanline}
\end{figure}

Figure~\ref{fig:scanline} shows a scan-line plot of a prediction from the HDR test set, using the DHH method, along with the LDR input and HDR target predictions. The~LDR input is saturated and constant for most of the line for all three channels (white), while the target and predictions vary and have strong correlations. The~target and prediction values are scaled to be close to the LDR values for visualisation purposes. It is observed that the prediction of the blue channel is lower than the target relative to the prediction of the other two. This is in agreement with the image, since the prediction is more yellow along the scan-line whereas the target image is blue. The~tone is reproduced adequately and blends well with the~surroundings.

\unskip

\begin{figure}[H]
{\captionsetup{position=bottom,justification=centering}
    \begin{subfigure}[t]{1.0\linewidth}
        \includegraphics[width=1.0\linewidth]{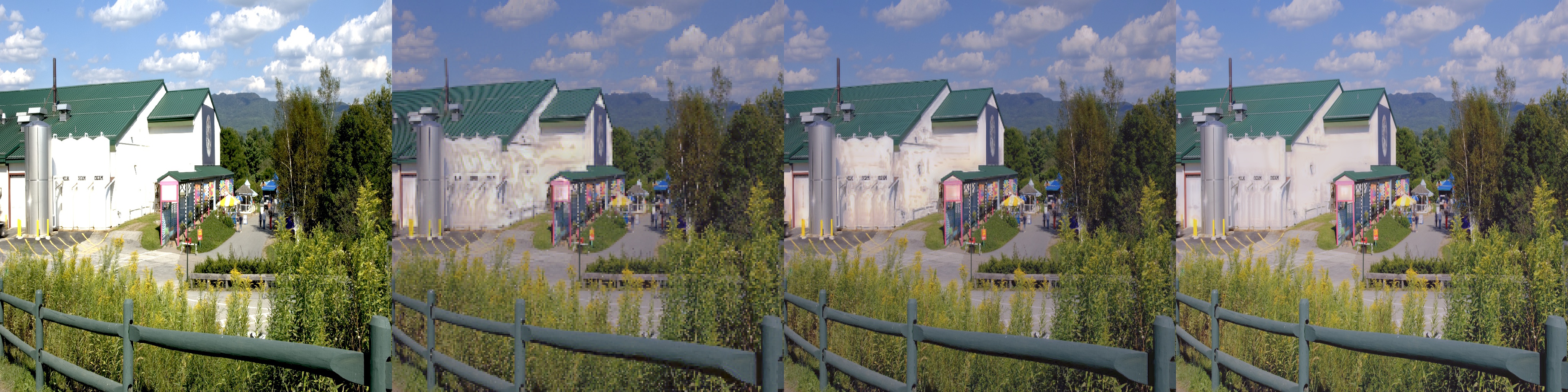}
        \caption{BenJerrys}
    \end{subfigure}
    \par\bigskip
    \begin{subfigure}[t]{1.0\linewidth}
        \includegraphics[width=1.0\linewidth]{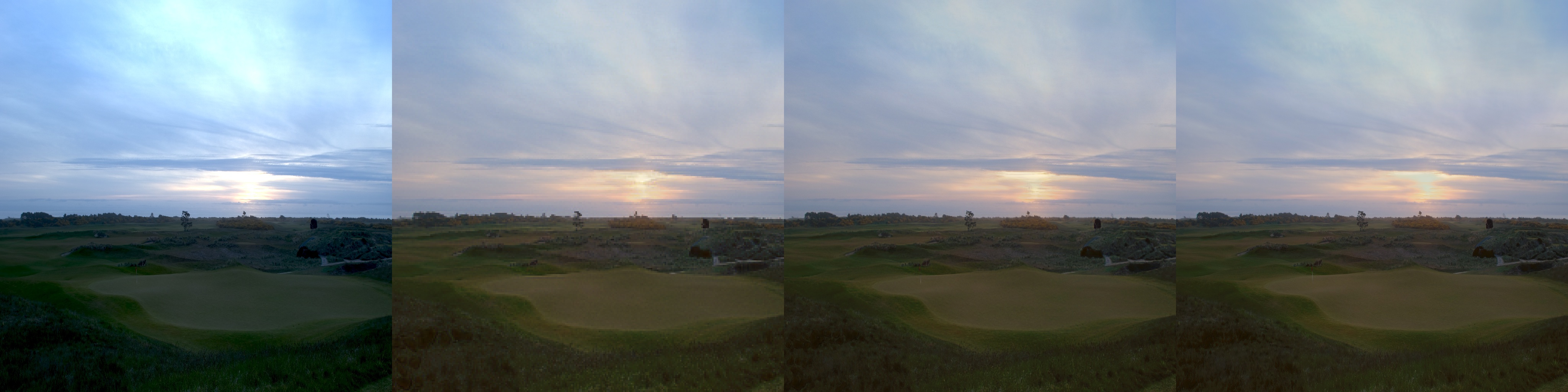}
        \caption{BandonSunset(1)}
    \end{subfigure}
    \par\bigskip
    \begin{subfigure}[t]{1.0\linewidth}
        \includegraphics[width=1.0\linewidth]{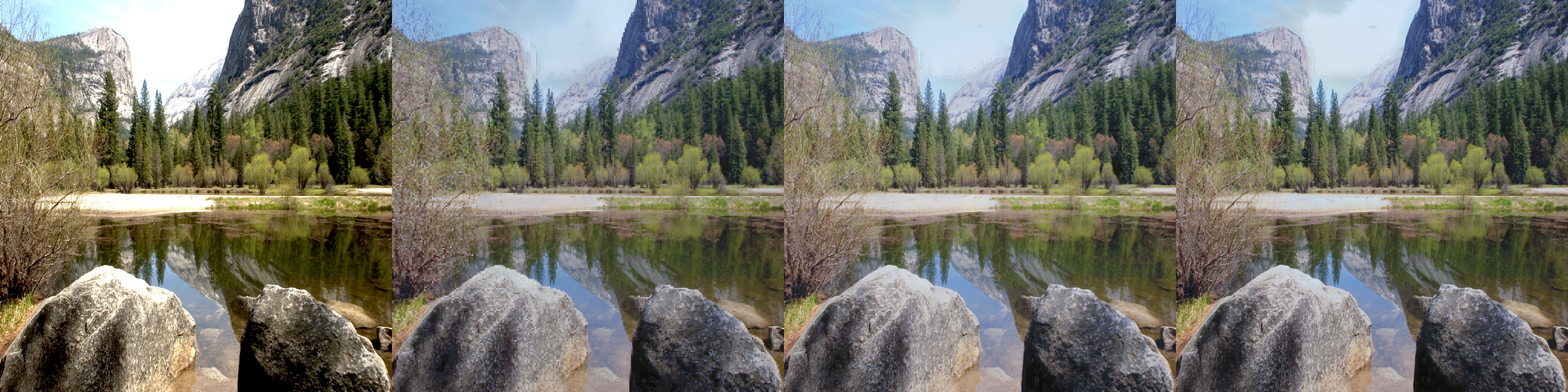}
        \caption{MirrorLake}
    \end{subfigure} }
    \caption{Test-set predictions for inputs at varying resolutions. First column is the
    input LDR, followed by predictions at resolutions of \(256 \times 256\), \(512 \times 512\)
    and \(1024 \times 1024\), respectively.}\label{fig:size}
\end{figure}

Figure~\ref{fig:size} shows test-set predictions using DHH at various resolutions. The~first column is the LDR input (culling) while columns 2--4 are (tone mapped) predictions from inputs of resolutions \(256 \times 256\), \(512 \times 512\) and \(1024 \times 1024\), respectively.  It is observed that the dynamic range expansion is consistent at different resolutions; however, the~hallucinations change structure depending on the input size. In~hard cases that are not common in the dataset, for~example the side of the building in Figure~\ref{fig:size}a, the~inpainting is not of good quality and the textures do not match the surroundings. The~clouds that are washed out in the input LDR are reconstructed well. Likewise, the~part of the sky around the sun which is saturated in Figure~\ref{fig:size}b, along with the grass in the bottom left which is under-exposed, is also reconstructed well. Same for the 
washed out sky in Figure~\ref{fig:size}c.

\subsection{Quantitative}\label{sec:quantitative}

While DHH is focused more on qualitative results and is not expected to outperform other methods quantitatively as metrics do not account for hallucinated content, the~results serve to demonstrate if any further quality in the rest of the image is being lost compared to the standard methods. In~addition, quantitative results serve as a validation for the use of the HDR pixel transform proposed in Section~\ref{sec:gunet:transform} and the hdr data augmentation regime from Section~\ref{sec:data_augmentation}.

\begin{figure}[H]
\widefigure
\begin{subfigure}[p]{0.78\linewidth}
    \begin{minipage}{0.47\linewidth}
        \includegraphics[width=1.0\linewidth]{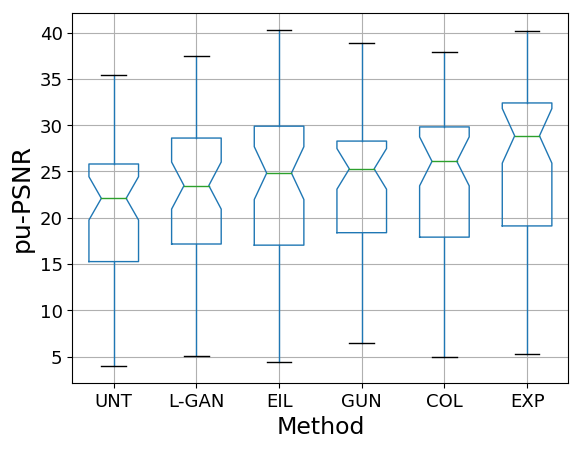}
    \end{minipage}
    \begin{minipage}{0.47\linewidth}
        \includegraphics[width=1.0\linewidth]{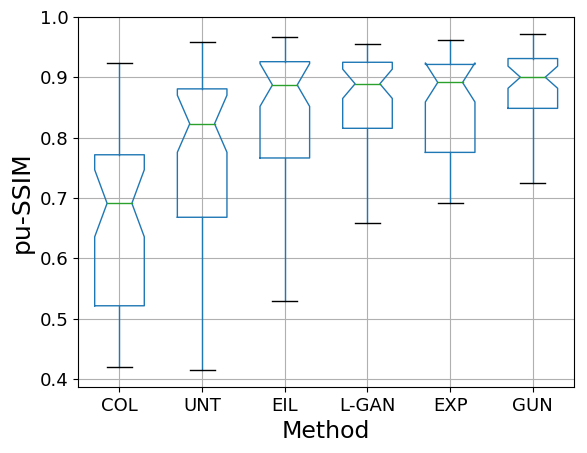}
    \end{minipage}\\
    \begin{minipage}{0.47\linewidth}
        \includegraphics[width=1.0\linewidth]{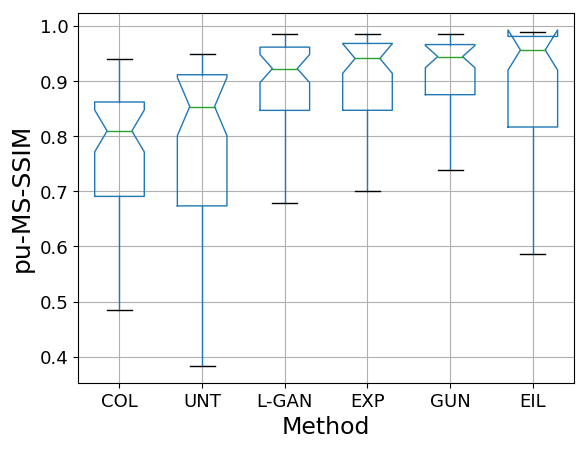}
    \end{minipage}
    \begin{minipage}{0.47\linewidth}
        \includegraphics[width=1.0\linewidth]{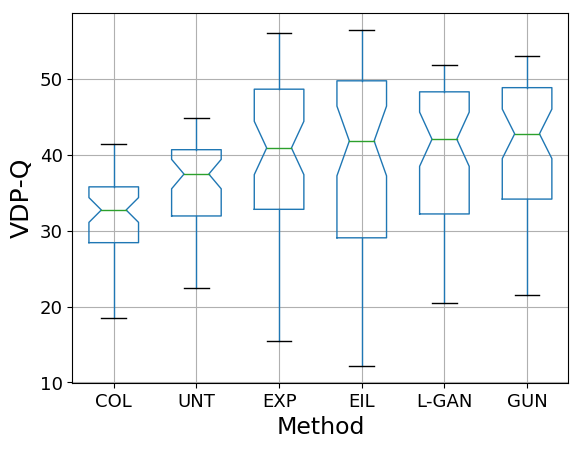}
    \end{minipage}
    \caption{{optimal}---scene-referred}\label{fig:halluc_boxplots_scene_optimal}
\end{subfigure}\\
\par\bigskip
\begin{subfigure}[htbp]{0.78\linewidth}
    \begin{minipage}{0.47\linewidth}
        \includegraphics[width=1.0\linewidth]{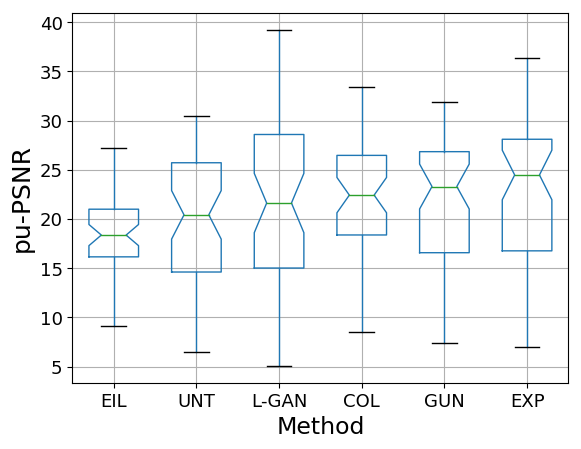}
    \end{minipage}
    \begin{minipage}{0.47\linewidth}
        \includegraphics[width=1.0\linewidth]{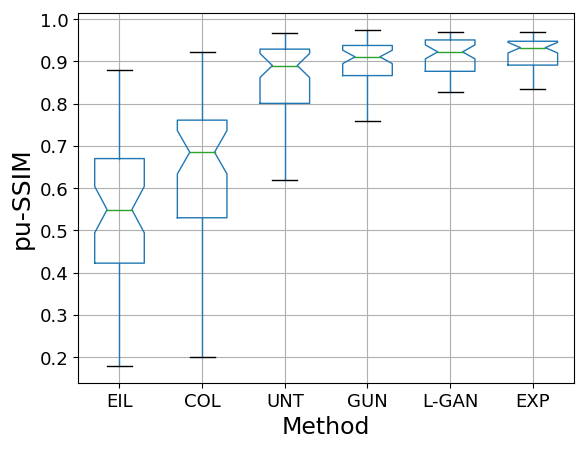}
    \end{minipage}\\
    \begin{minipage}{0.47\linewidth}
        \includegraphics[width=1.0\linewidth]{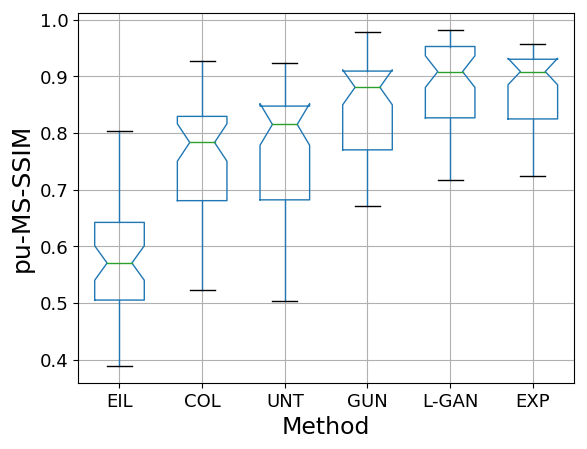}
    \end{minipage}
    \begin{minipage}{0.47\linewidth}
        \includegraphics[width=1.0\linewidth]{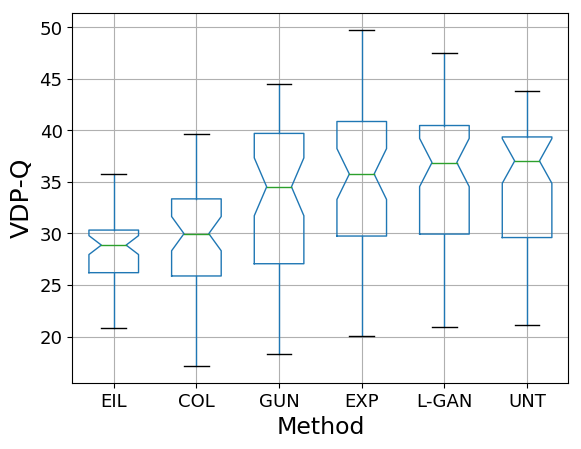}
    \end{minipage}
    \caption{culling---~scene-referred}\label{fig:halluc_boxplots_scene_culling}
\end{subfigure}\vspace{6pt}
\caption{Box plots of all metrics for scene-referred HDR obtained from LDR {culling} and {optimal} exposures. The~DHH method is referred to as L-GAN in the~plots.}
\label{fig7}
\end{figure}

\begin{figure}[H]
\widefigure
\begin{subfigure}[htbp]{0.78\linewidth}
    \begin{minipage}{0.47\linewidth}
        \includegraphics[width=1.0\linewidth]{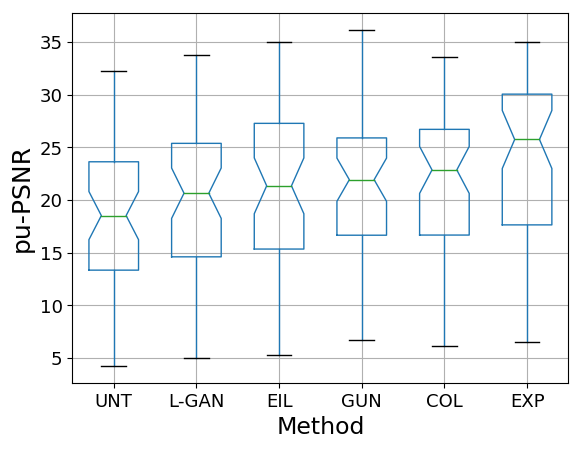}
    \end{minipage}
    \begin{minipage}{0.47\linewidth}
        \includegraphics[width=1.0\linewidth]{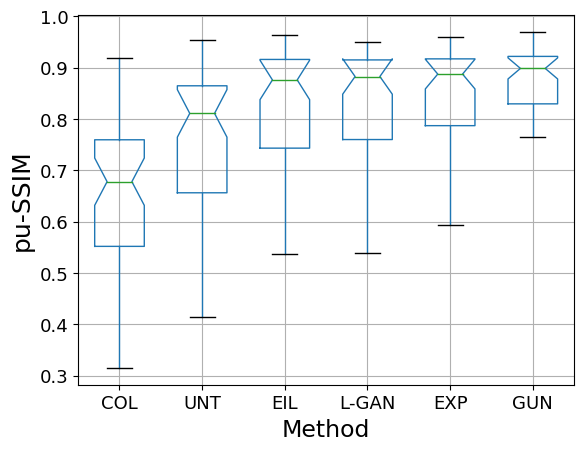}
    \end{minipage}\\
    \begin{minipage}{0.47\linewidth}
        \includegraphics[width=1.0\linewidth]{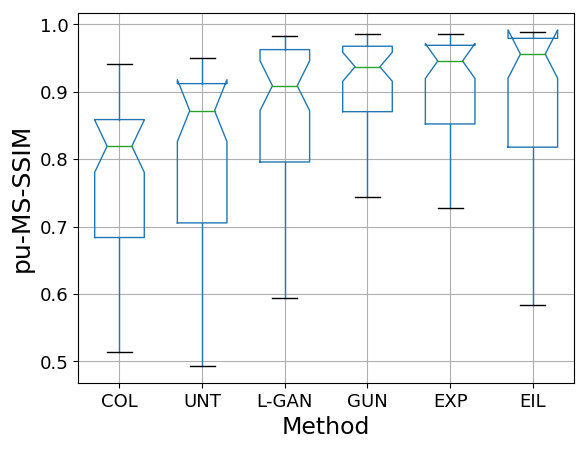}
    \end{minipage}
    \begin{minipage}{0.47\linewidth}
        \includegraphics[width=1.0\linewidth]{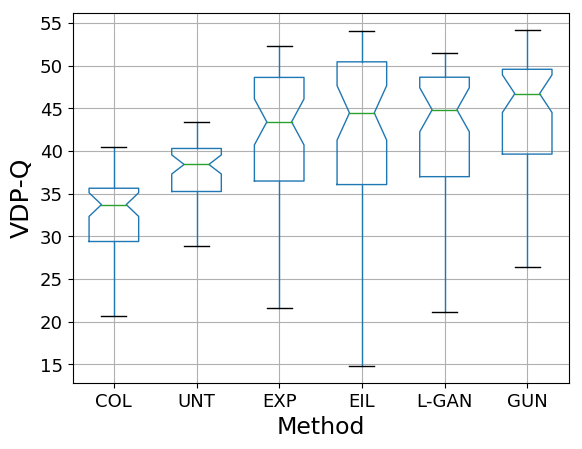}
    \end{minipage}
    \caption{{optimal}---display-referred}\label{fig:halluc_boxplots_tv_optimal}
\end{subfigure}\\
\par\bigskip
\begin{subfigure}[htbp]{0.78\linewidth}
    \begin{minipage}{0.47\linewidth}
        \includegraphics[width=1.0\linewidth]{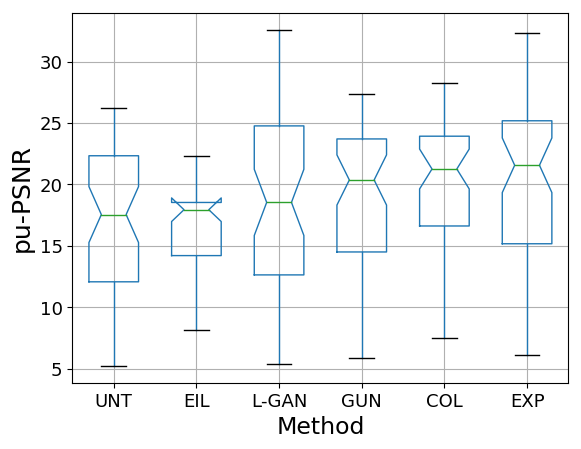}
    \end{minipage}
    \begin{minipage}{0.47\linewidth}
        \includegraphics[width=1.0\linewidth]{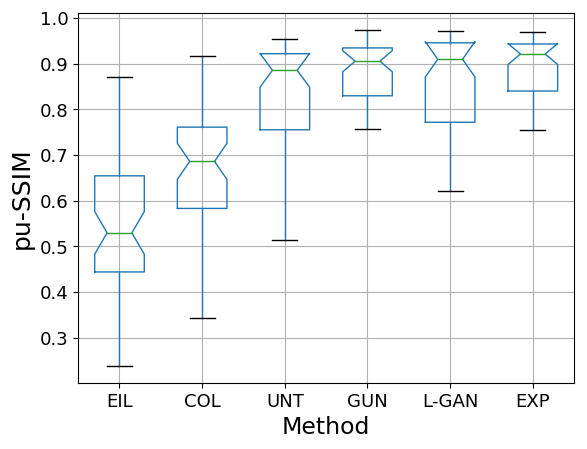}
    \end{minipage}\\
    \begin{minipage}{0.47\linewidth}
        \includegraphics[width=1.0\linewidth]{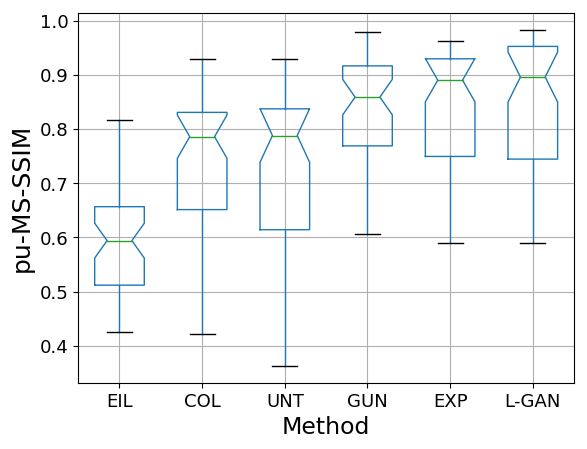}
    \end{minipage}
    \begin{minipage}{0.47\linewidth}
        \includegraphics[width=1.0\linewidth]{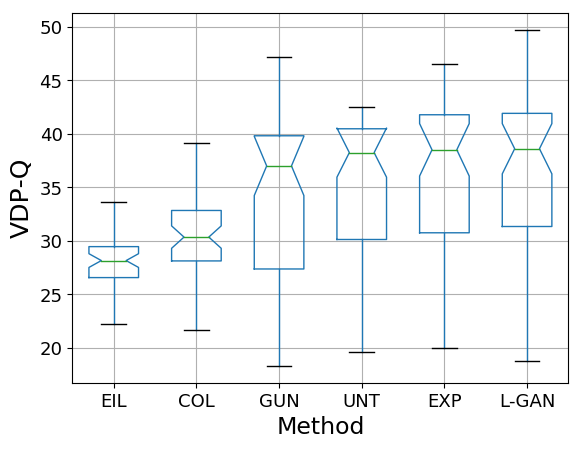}
    \end{minipage}
    \caption{{culling}---display-referred}\label{fig:halluc_boxplots_tv_culling}
\end{subfigure}\vspace{6pt}
\caption{Box plots of all metrics for display-referred HDR obtained from LDR {culling} and {optimal} exposures. The~DHH method is referred to as L-GAN in the~plots.}
\label{fig8}
\end{figure}

Figures \ref{fig7}a and \ref{fig8}b show quantitative results of the DHH method for the test-set compared with the CNN methods presented in ExpandNet~\cite{marnerides2018exp}, GUNet~\cite{marnerides2020spectrally} and the method by Eilertsen~et~al.~\cite{eilertsen2017cnn}, using the perceptually uniform (PU)~\cite{aydin2008extending} encoded PSNR, SSIM~\cite{wang2004image}, MS-SSIM~\cite{wang2003multiscale} and HDR-VDP-2~\cite{narwaria2015hdr}.

\clearpage
\end{paracol}
\nointerlineskip

\begin{figure}[H]
\widefigure
    \begin{subfigure}[t]{1.0\linewidth}
        \includegraphics[width=1.0\linewidth]{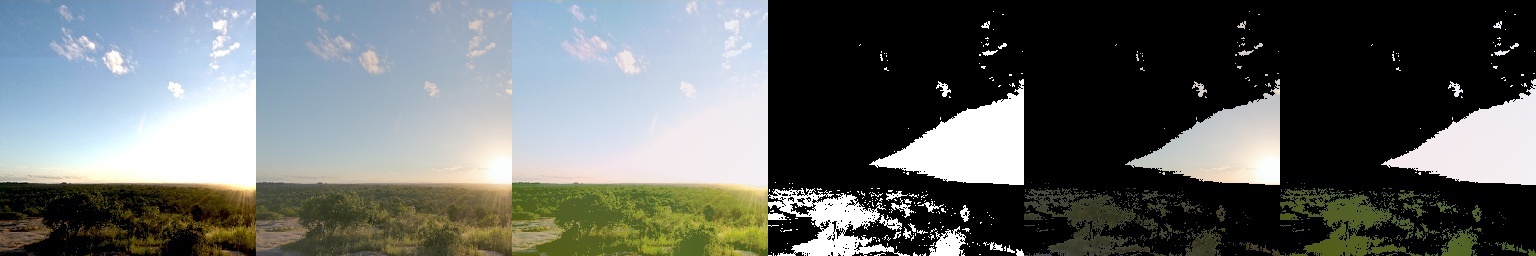}
    \end{subfigure}
    \begin{subfigure}[t]{1.0\linewidth}
        \includegraphics[width=1.0\linewidth]{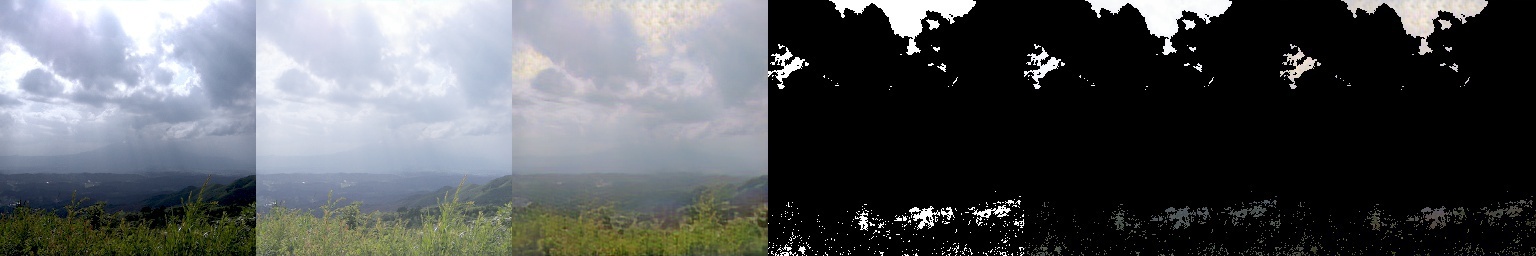}
    \end{subfigure}
    \caption{Training set samples for the GUNet and Autoencoder ablation studies. First column is the
    input LDR, followed by the target, prediction, mask, masked-target and masked prediction. Targets and predictions are tone-mapped. The~masks show the fully over/under exposed regions.}\label{fig:ablation_training}
\end{figure}

\begin{figure}[H]
\widefigure
{\captionsetup{position=bottom,justification=centering}
    \begin{subfigure}[t]{0.24\linewidth}
        \includegraphics[width=1.0\linewidth]{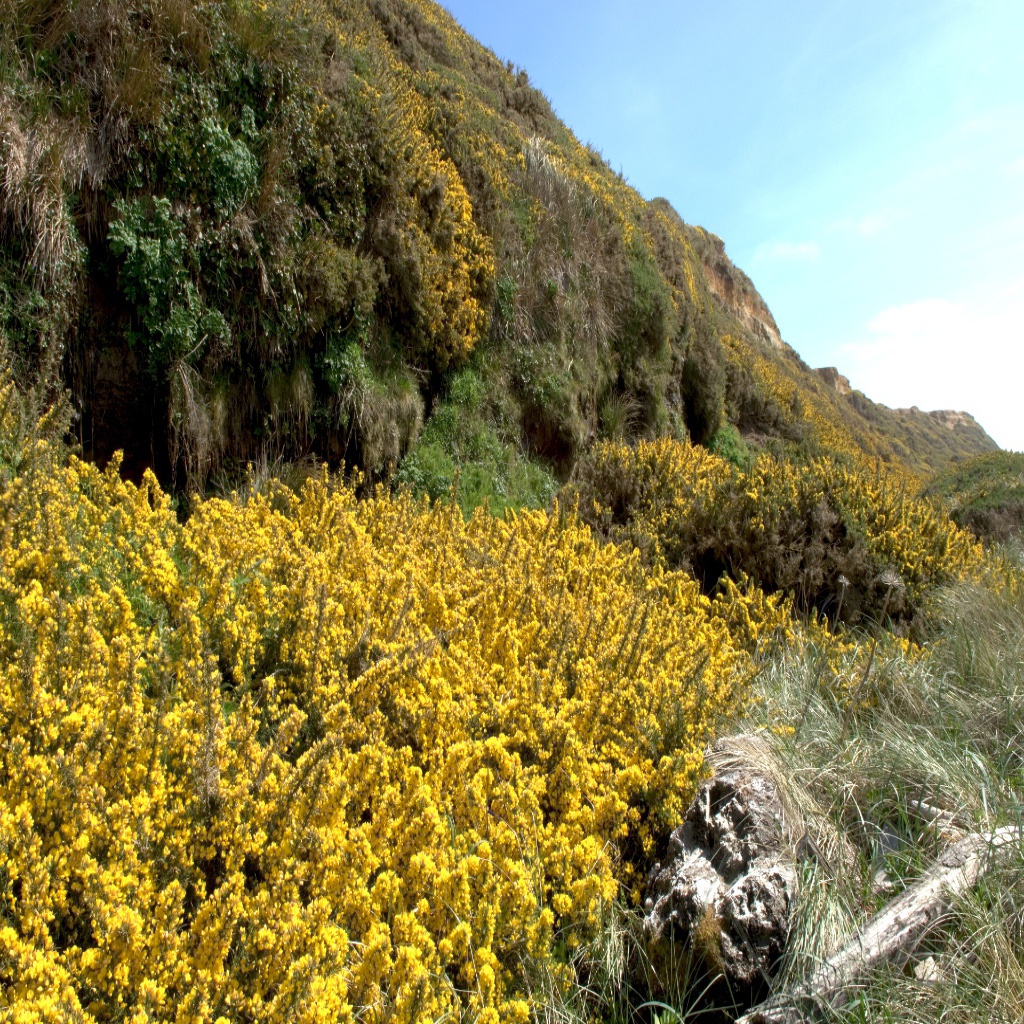}
        \caption{Input~LDR}
    \end{subfigure}
    \begin{subfigure}[t]{0.24\linewidth}
        \includegraphics[width=1.0\linewidth]{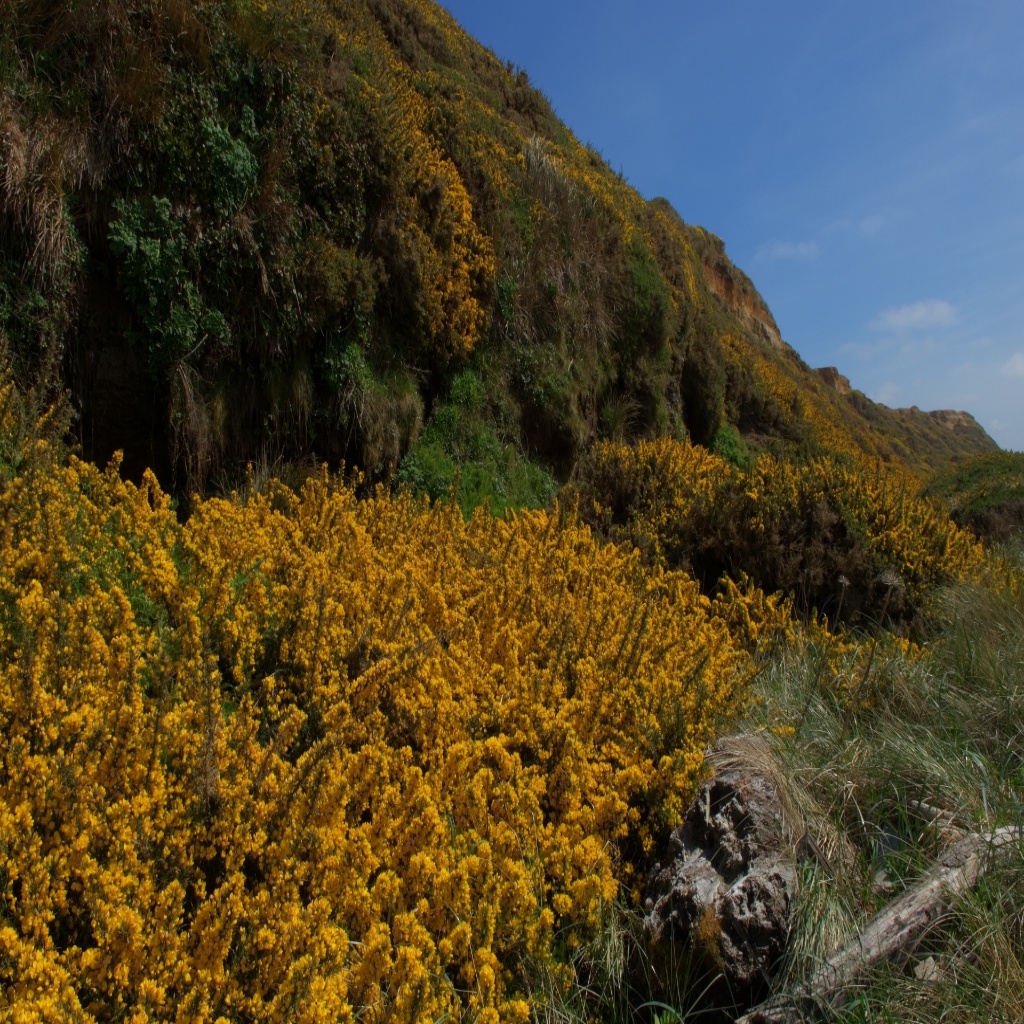}
        \caption{Target~HDR}
    \end{subfigure}
    \begin{subfigure}[t]{0.24\linewidth}
        \includegraphics[width=1.0\linewidth]{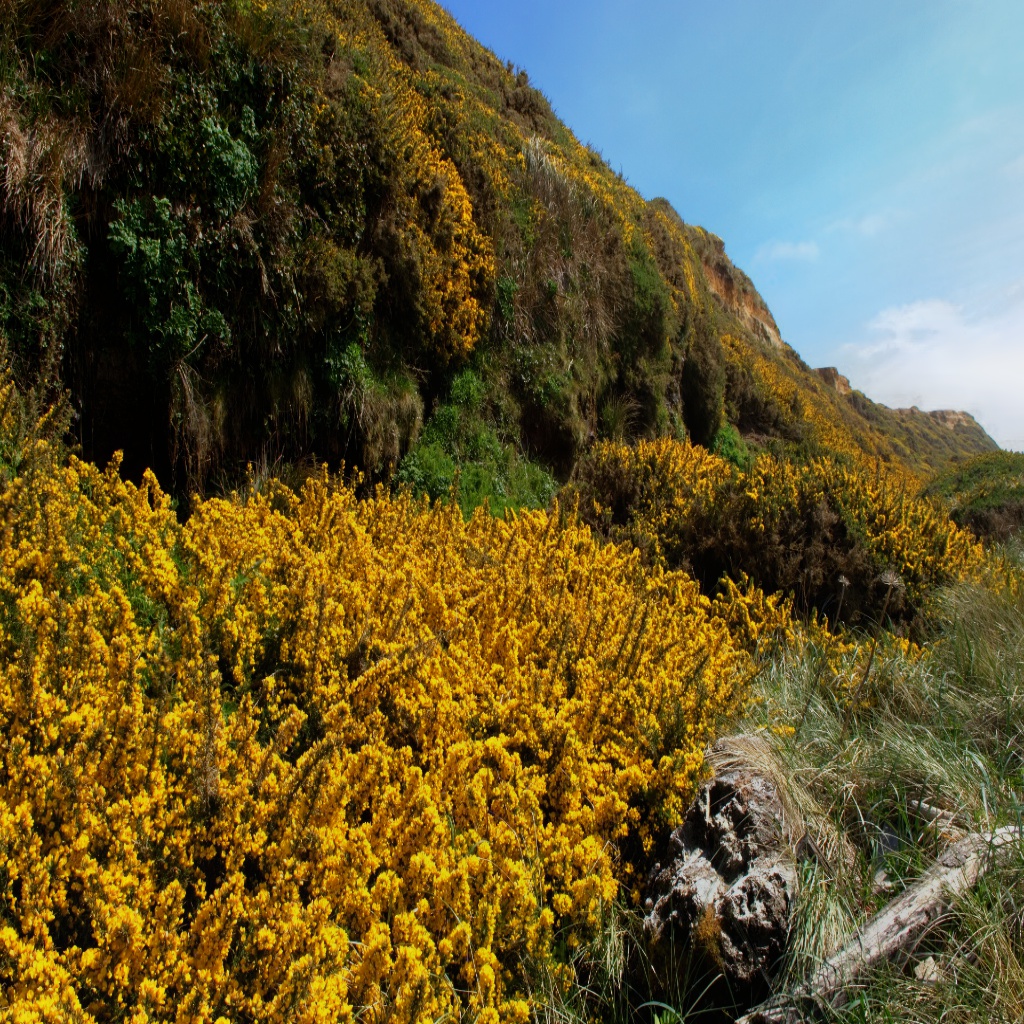}
        \caption{UNet + BN}
    \end{subfigure}
    \begin{subfigure}[t]{0.24\linewidth}
        \includegraphics[width=1.0\linewidth]{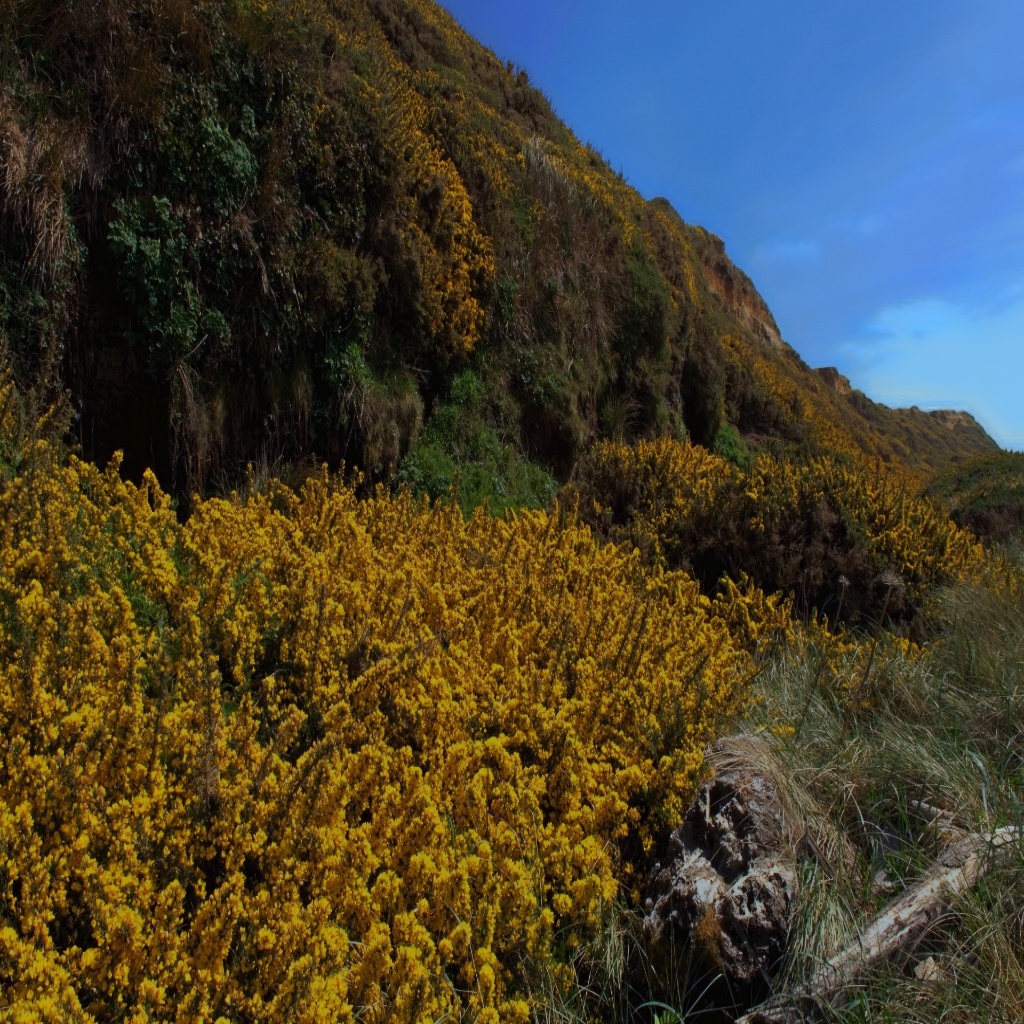}
        \caption{UNet + IN}
    \end{subfigure}}\vspace{6pt}
    \caption{Effect of batch normalisation and instance normalisation. Instance normalisation predictions are closer to the target HDR in~tone.}\label{fig:in_compare}
\end{figure}

\begin{paracol}{2}
\switchcolumn

The results for the DHH method are comparable with those of ExpandNet and GUNet and are not significantly different in the {culling} case.  While DHH does not dramatically improve on the quantitative performance of the other two methods, especially in the {optimal} exposure case where most of the content is well-exposed, these results are significant in multiple ways. First, they indicate that using a pre-trained network for data augmentation can produce good outcomes. The~DHH network was trained using data that was fundamentally LDR unlike the HDR data that the other networks were trained with. The~LDR data was expanded using the GUNet method and then tone mapped using exposure, to~remove content and use it as the network inputs. In~addition, the~loss used was based solely on GAN and perceptual losses that are CNN based, without~using standard regression losses,
which shows the effectiveness of these losses in comparison with more traditional ones. These results also demonstrate that even though hallucination is achieved to some degree, it is not at the cost of the dynamic range expansion of well-exposed~areas.

\subsection{Ablation~Studies}\label{sec:ablation}

Ablation studies were conducted to complement the results obtained using the DHH method. 
Alternative single-model end-to-end architectures are considered as well as the effects of the normalisation layer.
A GUNet~\cite{marnerides2020spectrally} architecture is considered, which was shown to perform well for dynamic range expansion of well-exposed areas, while keeping the performance of UNet architectures. An~additional Autoencoder network, sometimes termed a context encoder, which was previously used for LDR image inpainting~\cite{pathakcontext} is trained and compared.
A UNet architecture is also considered, similar to DHH, but~with Batch normalisation~\cite{ioffe2015batchnorm}, instead of the Instance Normalisation layers used in DHH. All the networks of the ablation studies similar in size. The~main encoder/decoder parts use features with sizes 64-128-256-512. The~bottlenecks are of size 512 and 8 layers deep, with~layers 2--7 using dilated convolutions of dilation size 2-4-8-8-8-8-8, respectively. All networks use residual modules, the~SELU activation~\cite{klambauer2017selu} and \(4 \times 4\) kernels for all layers except the Pre-Skip and Post-Fuse modules which use \(1 \times 1\) kernels. All UNet networks use bilinear downsampling and upsampling which was shown~\cite{marnerides2020spectrally} to have the least distortions in the spectrum compared to other non-guided UNet configurations.  The~Guided filter parameters for the GUNet architectures are the same as in the original~\cite{marnerides2020spectrally}. An~additional version of the DHH method is trained for consistency, using the above model configuration, which was chosen to be smaller to save computational resources, similarly to the other models in the ablation~studies.

Figure~\ref{fig:ablation_training} also presents samples for the GUNet and Autoencoder methods. 
The GUNet architecture successfully expands the well exposed areas and also attempts to hallucinate fully saturated and under-exposed areas. However the hallucinations are predominantly flat without any texture variations. This is an expected outcome due to the nature of the GUNet which has a strong guidance from the input features.
The autoencoder network provides hallucinations of varying textures in the badly exposed areas; however, it distorts the well exposed areas, adding defects of repeated patterns. The~lack of faithful reproduction of the well exposed areas can be attributed to the lack of skip~connections.

Figure~\ref{fig:in_compare} compares the effects of using batch normalisation and instance normalisation. Batch normalisation tends to ``average out'' the hallucination predictions, creating outputs close to the LDR input in tone. For~example, the~saturated part of the sky is more grey/white in the BNH prediction rather than the INH where it's blue. The~overall tone of the INH prediction is closer to the target HDR image, compared to that of BNH. This suggests that using instance normalisation provides improved hallucination~predictions.
\clearpage
\end{paracol}
\nointerlineskip

\unskip

\begin{paracol}{2}
\switchcolumn

\vspace{-6pt}

\section{Conclusions}\label{sec:conclusions}

This work presented DHH, a~GAN-based method, that expands the dynamic range of LDR images and fills in missing information in over-exposed and under-exposed areas. The~method is end-to-end using a single CNN, performing both tasks simultaneously. Due to the scarcity of HDR data, a~data augmentation regime is proposed for simulating LDR-HDR pairs for HDR hallucination, using a pre-trained CNN for dynamic range expansion. An~additional HDR pixel distribution transformation method is proposed, for~the normalisation of HDR content for use in deep learning methods.
Methods involving masking that combine an expansion and a hallucination network each focusing on different aspects of the problem were inspected. This approach produced visible boundary artefacts and tone mismatch for both grey and RGB linear masks. End-to-end methods using a single network provided improved results.
Networks with no skip connections, similar to context encoders previously used for other inpainting tasks produced noticeably more artefacts for well exposed areas and are may not be ideal for a complete end-to-end solution.
Normalisation layers have an effect on the global contrast of the prediction, giving more consistent predictions for larger inputs. Batch Normalisation can cause results to be more desaturated and closer to the LDR input, while Instance Normalisation that gives predictions closer to the target HDR\@. {Overall, the~results compare well with state-of-the-art methods}. 
Improvements can be directed towards the application of models on larger sized images with larger areas to hallucinate. In~addition, the~models can be adjusted such that they are able to better reproduce semantic details and not be limited to matching textures. {Future work will look into the application of the method to HDR video, ensuring temporal consistency across frames. Other applications that may be considered include colourisation, de-noising and single image superresolution along with inverse tone-mapping.}

\vspace{6pt} 
\authorcontributions{Conceptualization, D.M., T.B.-R. and K.D.; Methodology, D.M., T.B.-R. and K.D.; Software, D.M.; Supervision, T.B.-R. and K.D.; Validation, D.M.; Visualization, Demetris Marnerides; Writing---original draft, D.M.; Writing---review \& editing, D.M., T.B.-R. and K.D. All authors have read and agreed to the published version of the manuscript.}
\funding{This research received no external funding}




\dataavailability{Training data is available on \url{www.flickr.com}. Testing data is available on \url{http://markfairchild.org/HDR.html}. Both last accessed on 10/06/21.}
\conflictsofinterest{The authors declare no conflict of~interest. 
} 
 \end{paracol}
 \reftitle{References}

\end{document}